\documentclass{bmcart} 
\usepackage[utf8]{inputenc} %unicode support
\usepackage{amsmath}
\usepackage{graphicx} %package to manage images
\graphicspath{ {images/} }
\usepackage{amssymb}

\startlocaldefs
\endlocaldefs

\begin{document}

%%% Start of article front matter
\begin{frontmatter}

\begin{fmbox}
\dochead{Manuscript}
\title{Facial Information Recovery from Heavily Damaged Images using Generative Adversarial Network- PART 1}

\author[
   addressref={aff1},                   % id's of addresses, e.g. {aff1,aff2}
   corref={aff1},                       % id of corresponding address, if any
   noteref={n1},                        % id's of article notes, if any
   email={pushparajam@xrvision.com}   % email address
]{\fnm{Pushparaja} \snm{Murugan}}

\address[id=aff1]{%                           % unique id
  \orgname{XRVision Research and Development center, Singapore}, % university, etc                                  
}
\address[id=aff2]{%
  \orgname{XRVision technologies},
  \street{D\"{u} 71 Ayer Rajah Crescent},
  \postcode{639798}
  \city{Singapore},
  \cny{Singapore}
}

\begin{artnotes}
%\note{Sample of title note}     % note to the article
\note[id=n1]{} % note, connected to author
\end{artnotes}

\end{fmbox}% comment this for two column layout

\begin{abstractbox}
\begin{abstract} Over the past decades, large number of techniques have emerged in modern imaging systems to capture the exact information of the original scene regardless of shake, motion, lighting conditions and etc., These developments have progressively addressed the acquisition of images in high speed and high resolutions.  However, the various ineradicable real time factors cause the degradation of the information and the quality of the acquired images. The available techniques are not intelligence enough to generalize this complex phenomena. Hence, it is necessary to develop an intelligence framework to recover the possible information presented in the original scene. In this article, we propose a kernel free framework based on conditional-GAN  to recover the information from the heavily damaged images. The degradation of images are assumed to be occurred by combination of various blur. Learning parameter of the cGAN is optimized by multi-component loss function that includes improved wasserstein loss with regression loss function. The generator module of this network is developed by using U-Net architecture with local Residual connections and global skip connection. Local connections and a global skip connection are implemented to reutilization of all stages of features. Generated images shows that the network have the potential of recover the probable information of blurred images from the learned features.  This research work is carried out as a part of our IOP studio software 'Facial recognition module'.

\end{abstract}


\end{abstractbox}
%
%\end{fmbox}% uncomment this for twcolumn layout

\end{frontmatter}

\section*{Introduction}

Images are representation of visual information in digital form but the image acquisition and formation process degrade the information of the representation of the original scene while capturing it. Blur, point-wise non-linearities and the noise formation are common case of degradation that usually occurs by the image sensing system. Image blur is a unavoidable information degradation. In another hand, it is form of bandwidth reduction of the images due to the image formation process. One of the easiest possible solution is to capture the images in shorter exposure intervals during the image acquisition process. In this case, noise formation in capture is inevitable when capturing the images in dark lighting conditions. Another possible solution is information recovery and reconstruction / off-line process. There has been a lot of developments in digital image processing techniques to sustain the visual informations and to increase the quality of the images while capturing the images in off-line . The main objective of the information recovery and restoration is to estimating the possible information of an original scene from the degraded images. It severs in many area such as astronomy, medical image process and satellite image processing as well as the commercial photographic industries. Information recovery and reconstruction can be carried by image in-printing and deblurring process \cite{puetter2005digital} \cite{almeida2010blind} \cite{gunturk2012image} \cite{ji2012robust}.
	Image in-painting is a process of generating possible information to fill the damaged or missing regions in an image by utilizing the available information that includes restoration of old images, removing scratches, texts, special effects and filling the damaged regions \cite{bertalmio2003simultaneous}. The in-printing process can be largely classified into two categories such as structural in-printing and texture in-printing. Structural in-printing is concerned with propagation of structure into the missing region and synthesization of texture on that area which can be very effective for in-printing small region. The texture in-printing is using the global information from multiple images and filling the missing regions which is very effective in large missing area \cite{kang2002inpainting}. In early 1990's, first image in-paint model is proposed based on non-linear partial differential equations to restore the information in the damaged images. In this method, the gray level information is propagated in the direction of isophotes to obtain a full image \cite{bertalmio2000image}. Bertozzi is proposed a method for in-printing image through two dimensional fluid dynamics navier-strokes equation \cite{bertalmio2001navier}. Cheng-Shian  Lin  and  Jin-Jang  Leou  suggested  an another  four-step approach  for  inpainting \cite{shen2002mathematical}.  Marcelo Bertalm et al proposed a hybrid approach of using structure and texture in-printing. The idea of this approach is to decomposing the images into two functions by their character and work on those functions with structure and texture filling \cite{rane2003structure}. 
    
    Information degradation caused by the blur effects produces visually unattractive images. Fast moving objects, image acquisition in dim lighting condition, capturing long distance objects, other side of focused area in the images are typical example for the blur generation where even the high speed and higher resolution sensing system perform very badly \cite{wang2014recent}.  Image deblurring is an inverse problems where the reconstruction or recover the information of shared images from the degraded images \cite{tikhonov1977vy}. Numerous investigation is carried out to deblur the images based on non-blind deblur and blind deblur method. In non-blind deblur methods such as Richardson-Lucy \cite{richardson1972bayesian} and  Wiener filter \cite{brown1992introduction} , the blur kernal is assumed to be known and the information recovery is carried out by using the blurred images and the  kernels.  In blind deblur methods,  the kernel is known to be unknown and the estimation of kernel is carried out by using the blurred images.  However, the recent developments in deblur methods are presented to tackle the both blind and non-blind problems. Due to the ill-nature of the blur kernels as well as the noise in the kernel, the acquired blurred images doesn't exactly represent the information about the original scene and true kernels are significantly mismatched \cite{bertero1998introduction}.

	The recent development in convolutional Neural Network (CNN) would provide the possibility of addressing this limitation. The architecture of CNN, learning of CNN, hyperparameter optimization, and the limitation are elaborately discussed in our previous studies \cite{murugan2017feed} \cite{murugan2017regularization} \cite{murugan2017hyperparameters} \cite{murugan2018implementation}. Another remarkable invention based on CNN architecture is Generative Adversarial Network. A typical CNN architecture has two convolutional neural networks. Image style transformation, deblur GAN, text to images are the one of most promising developments in translation and transformation of image information. Generative Adversarial Network is an artificial intelligence technique, consists of two networks namely discriminator and generator competing with each other in zero sum game. The idea of this generative model is firstly introduced by Ion Goodfellow \cite{goodfellow2014generative}. The generator networks learn to map from the latent space and generates the image from the data distribution while the discriminator networks discriminates the real data distribution and the generator data distribution. The training objective of the GAN network is to increase the error rate of the discriminative network.  The subsequent development of GAN network such as DCGAN, improved DCGAN improved the performance of the vanilla GAN. In DCGAN, the author suggested that the importance of Batch Normalization in both generator and discriminator module as well as the importance of avoiding fully connected layers and striding instead of pooling \cite{radford2015unsupervised}.  The technique proposed in improved DCGAN allows to generator high resolution images  where the authors has suggested various enhancements on training such as feature matching, Historical averaging, One-sided label smoothing, and Virtual batch normalization \cite{salimans2016improved}.  Another  GAN network is conditional GAN, known as CGAN where utilization label information is resulted with better quality of image generation and governance over the images.  One of the important invention in GAN is Wasserstein GAN which overcomes the limitation of vanilla GAN by optimizing the learning parameters by using wasserstein Earth Mover distance as the objective function \cite{arjovsky2017wasserstein}.  Super resolution GAN \cite{ledig2017photo}, pix2pix GAN, cycle GAN  \cite{isola2017image} are the notable promising inventions in generative models that provide the possibilities of translation of image information from the real data distribution  over a noisy data distribution.  In super resolution GAN, the authors proposed an content loss objective functions along with the adversarial loss function. Content loss function is Euclidean distance between the high level feature of generated and real data image distribution which allows the generation of more similar images to the high resolution original image.  Yeh et al proposed an architecture for image in-printing to fill the information in the mission region of the images \cite{yeh2016semantic}.  Ramakrishnan et al, proposed a kernel free blind algorithm to deblur images by using pix2pix and fully connected dense layer \cite{ramakrishnan2017deep}. From the recent invention in Generative Adversarial Network, it is been clearly understood that the GAN network have the potential of preserving the inherent textural and structural information and generating convening images  that looks more close to the real data distribution.

\section{Proposed Network}

\begin{figure}[h]
         \includegraphics[width=12cm]{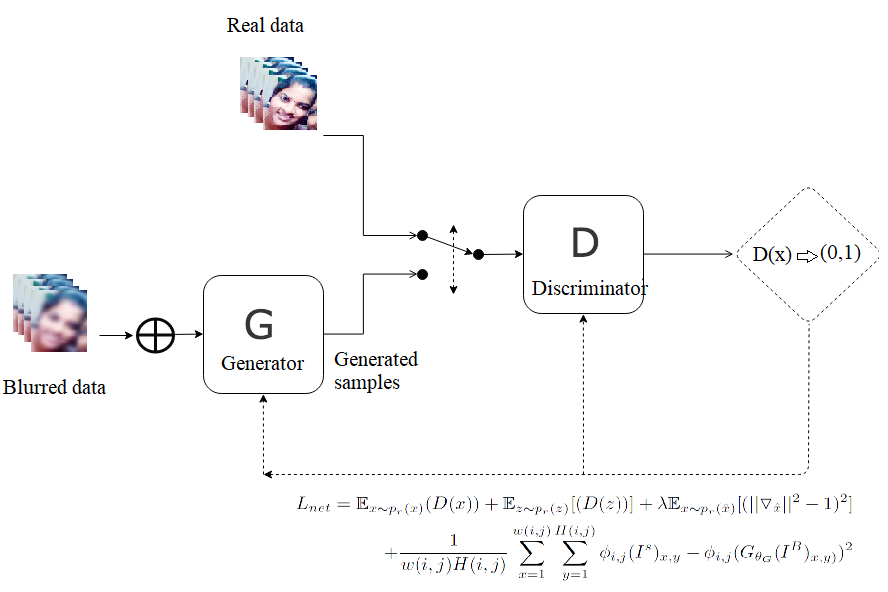} \label{gan}
         \caption{GAN architecture}
\end{figure}

Recently, Generative Adversarial Networks are finding important role in supervised, semi-supervised as well as unsupervised learning vision tasks as the generative models implicitly learn probability density of high dimensional distributions of the data and generate natural looking images. The generator and the discriminator in the GAN network competing each other in zero sum game to optimize the learning parameters.  The schemantic of the GAN network is shown in the Figure. \ref{gan}. The generator generate images of natural looking data samples from noise input data to fake the discriminator while the discriminator tends to distinguish the generated samples from the read data. Both the forger (Generator) and the expert (Discriminator) learn simultaneously by minimize the distance between the probability distribution of real and generated data. However, while the discriminator has the access to the generated data and real data, the generator has no access to the real data distribution. The noise input data to the discriminator provide the possible information about the ground truth to distinguish between the synthetic  generated data and real data distribution. The same noise data distribution is used for training the generator to produce natural looking images close to the real data with superior quality. The generator and the discriminator composed of deep convolutional layer and fully connected dense layers. Since the necessity of direct invertible of the generator and the discriminator, the both network modules has to be continuous and differentiable everywhere. 

In a typical GAN architecture, the discriminator network $D$ maps the generated images distribution to the real data distribution $D:D(x_i) \rightarrow [0,1]$ and the generator learn to map the representations of the latent space to the space of data distribution $G \rightarrow \mathbb{R}^{|x|}$, where $z \in \mathbb{R}^{|x|}$ is represent the samples from the latent space $x \in \mathbb{R}^{|x|}$ of image distribution. For a fixed generator $G$, the discriminator $D$ is trained to classifying the images as the fake and real input.  Once the discriminator is trained optimally, the generator is continued to learn to generate images close the original images.  These captured statistical distribution of the training data is applied to solve wide variety of problems such as semantic image editing, style transfer, data augmentation and image retrieval, translation and transformation.  Detail overview of super resolution, style transfer, photo generation are discussed in \cite{isola2017image}.  Conditional GAN known as cGAN learns the statistical distribution from the training data and the noise vector $z$ to $y:G : x, z \rightarrow y$ by placing a condition on the discriminator. Markovian discriminator  allows to achieve perceptually superior results on generation of images from label maps reconstructing objects from the edge maps, and the colorizing images.

\subsection{Generator and Discriminator}
\begin{figure}[h]
         \includegraphics[width=13cm]{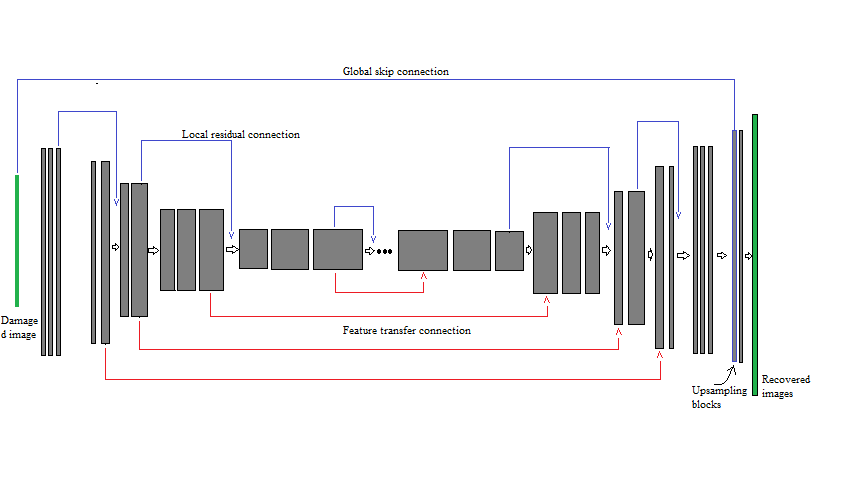} \label{generator}
         \caption{Generator architecture}
\end{figure}

The network generator architecture is shown in the Figure \ref{generator}. It consists of one convolutional block at the head, two convolutional block at the rear side, seven residual blocks. Residual blocks consist of four sequent convolutional layer, instance normalization layer and the activation function sequentially.  The output of the third activation layer in every residual blocks is internally connected with output of the first activation layer in next residual blocks. Along with these local connection, a global skip connection is also introduced.  Drop with probability of 50 \% is implemented in the residual blocks.  InstanceNorm and LeakyReLU with $\alpha = 0.1$ is introduced in every convolutional network expect the last layer.  Reutilization of features between the subsequent layers allow the network to reconstruct the possible information from the learned features.  Also, it is noted that the performance of the architecture is higher even with the smaller network. 

The discriminator in the GEN architecture is the expert that distinguish the difference between real data and generator data. In other hand, it helps the generator to generate more realistic information from the learned data distribution. In this architecture, Markovian patch discriminator with ten convolutional layer is implemented which also enforce the coloration in the generated natural images.  
\subsection{Loss function}

The objective of the Generator is to learn the distribution $p_{\theta}(x)$ , approximate to the real distribution $p_{r}(x)$ and generate samples such that the probability density function of the generated samples $p_G(x)$ equals to the probability density function of the real samples $p_r(x$).   This can be approached by for learn the differential function $p_{\theta}(x)$ such that $p_{\theta}(x) > 0$ and $\int_x p_{\theta}(x) dx =1$ directly and optimize through maximum likelihood or learn the differential transformation function $q_{\theta}(z)$ of $p_{\theta}(x)$ and optimize through maximum likelihood where $z$ is the existing common distribution such as uniform or Gaussian distribution. 

The discriminator $D$ has to recognize the data from the real data distribution $p_r(x)$, where $D$ indicates the estimated probability of data points $x_i \in \mathbb{R}^n $. In case of binary classification, if the estimated probability $D(x_i): \rightarrow \mathbb{R} ^n [0,1]$ is the positive class $p_i$ and $1- D(x_i): \rightarrow \mathbb{R}^[0,1]$ is the negative class $q_i$, the cross entropy distribution between $p_i$ and $q_i$ is, 
\begin{align}
L(p,q) = - \sum_i^n p_i \log q_i \label{entro}
\end{align}
For a given point $x_i$ and corresponding label $y_i$, the Eq \eqref{entro} can be expressed as, 
\begin{align}
L((x_i,y_i), D) = - y_i \log D(x_i) - (1-y_i) \log (1-D(x_i))
\end{align}
It is been understood from the above equation, one of term tends to set to zero depending on the values of $y_i$. For the entire dataset distribution, the above equation can be written as,
\begin{align}
L((x_i,y_i)_{i=1}^n, D) = - \sum_{i=1}^n y_i \log D(x_i) - \sum_{i=1}^n (1-y_i) \log (1-D(x_i)) \label{fin}
\end{align}
In Generative Adversarial Network, the data distribution $x_i$ can be from the real data $x_i \sim p_r(x)$ or the generator data $x_i \sim p_g(z)$. Addition to that, we expect exactly half of data from the two sources. In order to encode this information probabilistically in the Eq \eqref{fin}, the sum is replaced with expectation $\mathbb{E}$ and label $y_i$ is with half of the values.  Hence, the loss function is,
\begin{align}
L((x_i,y_i)_{i=1}^n, D) = - \frac{1}{2} \mathbb{E}_{x \sim p_r(x)} \log(D(x)) -  \frac{1}{2} \mathbb{E}_{z \sim p_r(z)} \log[1 - (D(G(z)))]  \label{eq:4}
\end{align}In optimizing the value $L((x_i,y_i)_{i=1}^n, D)$, for the given real data distribution $p_r(x)$,  the estimated probability $D$ over the real data accurate by maximizing the value $ \mathbb{E}_{x \sim p_r(x)} \log(D(x))$ and  for given fake distribution from the generator$p_g(x)$ is close to zero by maximizing the value $ \mathbb{E}_{z \sim p_r(z)} \log[1 - (D(G(z)))]$.  On other hand, the generator is trained to increase the chance of producing the estimated probability high for the fake data by mini zing the $ \mathbb{E}_{z \sim p_r(z)} \log[1 - (D(G(z)))]$.  Hence, the generator and discriminator tends to fight each other in minmax game to minimize the loss function. 
\begin{align}
\min_G \max_D L(G, D) = - \frac{1}{2} \mathbb{E}_{x \sim p_r(x)} \log(D(x)) -  \frac{1}{2} \mathbb{E}_{z \sim p_r(z)} \log[1 - (D(G(z)))]  \label{eq:5}
\end{align}If the discriminator is trained before the generator parameter $G_{\theta}$ update, the minimization of loss function is equal to minimization of Jensen-Shannon divergence between the real data distribution $p_r(x)$ and generated data distribution $p_g(x)$. However, optimizing the value function Eq. \eqref{eq:5} suffer from vanishing gradient and model collapse as the discriminator saturates on GD.  They proposed an another method to measure the probability distribution based on Wasserstein distance, known as Earth Mover distance. It can be stated as, minimum transportation cost of moving the mass from the distribution $q$ into the distribution $q$, provided the $p$ and $q$ distribution are continuous and differentiable everywhere. The Wasserstein loss function can be expressed as by using Kantorovich-Rubinstein duality , 
 
\begin{align}
\min_G \max_{D \in \mathcal{D}}  L(G, D) = \mathbb{E}_{x \sim p_r(x)} (D(x)) + \mathbb{E}_{z \sim p_r(z)} [(D(z))]  \label{eq:6}
\end{align}
where, $\mathcal{D}$ is the set of 1-Lipschitz functions . After the optimization of the network , to ensure the discriminator probability estimation is to be in the set of 1-Lipschitz functions, the author introduce the weight clipping though it lead to undesirable results. In order to overcome this, gradient penalty term is introduced, 
 \begin{align}
  L_{adver}(G, D) = \mathbb{E}_{x \sim p_r(x)} (D(x)) + \mathbb{E}_{z \sim p_r(z)} [(D(z))] + \lambda    \mathbb{E}_{x \sim p_r(\hat{x})} [(||\triangledown_{\hat{x}} ||^2 -1 )^2] \label{adver}
\end{align}

As stated in the original paper, $\lambda$ is kept as 10 during the learning.  When gradient penalty in the Eq \eqref{adver} is fully optimized by back propagation, the discriminator will act as 1-Lipschitz function. However, if the 'image latent information transforming GAN architecture' is trained without the perceptual content loss network don't converge to meaningful state. Perceptional loss is nothing but the $L_2$ loss and  can be defined as Euclidean difference between deep network feature maps of the real data samples and generated data samples.  The perceptional loss is given in the Eq. \eqref{eq.prec}. 
\begin{align}
L_{percep} = \frac{1}{w(i,j)h(i,j)} \sum_{x =1}^{w(i,j)} \sum_{y=1}^{h(i,j)} \phi_{i,j} (I^s)_{x,y}  - \phi_{i,j} (G_{\theta_G}(I^B)_{x,y)})^2 \label{eq.prec}
\end{align}where,  $\phi_{(i,j)}$ is represent the $j^{th}$ feature maps of deep Convolutional Neural Network before $i^{th}$ max-pooling layer.   $w_{(i,j)}$ and $h_{(i,j)}$ are pixels dimensions of the feature maps.  The adversarial network loss function is combination of the above loss functions \eqref{adver} \eqref{eq.prec}. Hence the total loss is given by,
\begin{align*}
L_{net} =  \mathbb{E}_{x \sim p_r(x)} (D(x)) + \mathbb{E}_{z \sim p_r(z)} [(D(z))] + \lambda    \mathbb{E}_{x \sim p_r(\hat{x})} [(||\triangledown_{\hat{x}} ||^2 -1 )^2]  \\ + \frac{1}{w(i,j)H(i,j)} \sum_{x =1}^{w(i,j)} \sum_{y=1}^{H(i,j)} \phi_{i,j} (I^s)_{x,y}  - \phi_{i,j} (G_{\theta_G}(I^B)_{x,y)})^2
\end{align*}

\section{Dataset preparation }

The real blurs in the images are extremely complex which cannot be approximated simple parametric model to generate synthetically. Also ,It is very unlikely to happen to acquire the image pairs of blurred image and corresponding shape images to train the GAN network.  Hence, the image pairs are created artificially.  The high resolution sharp images are collected from various mobile phone camera images. YOLO network is trained to localize the faces in the images. The cropped images are later used in the training.  The main objective of the generating blurred images is to degrade the information presented in the original image. In that connection, high degree of motion blur, camera shake blur and defocus blur is applied to the original image.  There are $60,000$ image pairs are created for training. Since we want only learn the statistical information about the sharp image, the repeated image is also present in the training but with different blur.  The original image and the corresponding blurred images are taken for training the GAN network.  Many investigation is also successfully developed to develop synthetic blur images. These method proposed that the synthetic blurred images can be created by  convolving the shape images with linear motion of blur kernel or randomly sampling sie random points and fit a spline. In our synthetic generation of blur images are concerned with varying the direction of the blur kernel.  
\subsection{Image blur generation}
\begin{figure}[h]
         \includegraphics[width=12cm]{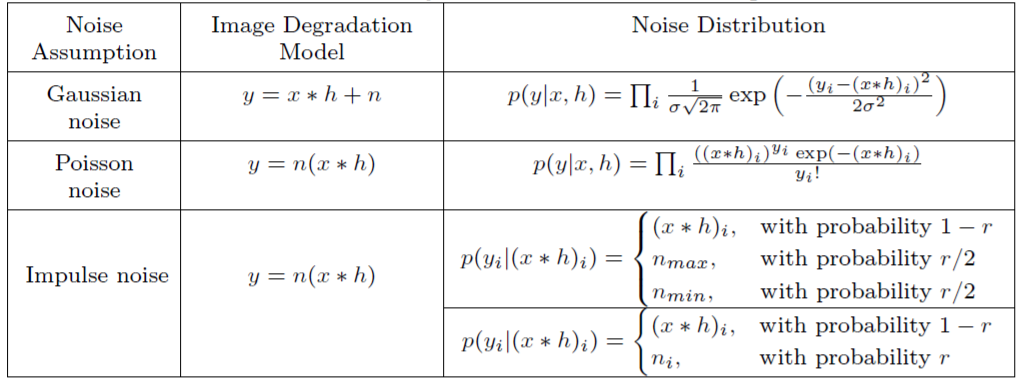} \label{noise}
         \caption{Degradation model based on the noise assumption}
\end{figure}

The blur kernel, known as point spread functions causes the image pixel to record light photons from the multiple scene points. In real time, many factor can cause the image blur that can degrade the information and quality of the objects appeared in the scene \cite{afonso2010fast}. Commonly image blur can be induced by object motion,  atmospheric turbulence, physical intrinsic, camera shake and defocus.  In classical deblur algorithm methods, the information recovery from the degraded images requires the understanding the kernel and appropriate modeling of the information presented.  Also, it highly complicated to generate blur images that could occur in real time. Hence, its necessity to understand the image formation models.  Image formation posses the information of radiometric and geometric by projecting the 3-D world in to the 2-D focal plane. The light rays passes through the camera lens is projected into the focal points. This can be modeled as, concatenation of the perspective projection and geometric distortion. 
The digital information of the images are formed by the discretization of the analog images which is transformed by the light photons \cite{delbracio2012non}. This can be expressed as,

\begin{align}
y_{blur} = S(f(D(P(s_r)*h_{ex}*h_{in})) + n_i \label{eq:blur}
\end{align}
where, $y_{blur}$ is represent the absorbed blurred image as the function of sampling operator, $P(s_r)$ is represent the perspective projection of the real planer scene,  $h_{ex}$ is represent the extrinsic kernel blur, $h_{in}$ is represent the intrinsic kernel blur and $*$ is represent the convolution operation.  The \eqref{eq:blur} is show the blur image formulation. The information recovery from the blurred image can expressed as ignoring the sampling effects,
\begin{align}
y_{blur} =  f(x * h) + n
\end{align}
where, $x$ is the latent sharp image from the $D(P(s_r)$ and h is the estimated kernel by combining the extrinsic$h_{ex}$ and intrinsic blur $h_{in}$. The general objective is to recover the information $x$ and the kernel $h$ from the degraded absorbed blurred images $y_{blur}$. For simplification, the effect of camera response function can be neglected and the blur  generation can be written as, 
\begin{align}
y_{blur} =  x * h+ n
\end{align}
Considering the whole image, this can be expressed with matrix-vector form as follows,
\begin{align}
y_{blur} =  Hx+ n
 \end{align} 
$n$ is the noise happen to be appear in the information of the image by the sensing system. $'n'$ cab be modeled as Gaussian noise, Poisson noise, and impulse noise. This various degradation models based on the noise assumptions is shown in the Figure \ref{noise}. 
Though many other factors cause the blur in the images, $H$ can be characterized as a specific properties of the blur which leads to different types of blur such as motion blur, camera shake blur, defocus blur and atmospheric intrinsic blur.
\begin{figure}[!hpb]
         \includegraphics[width=12cm]{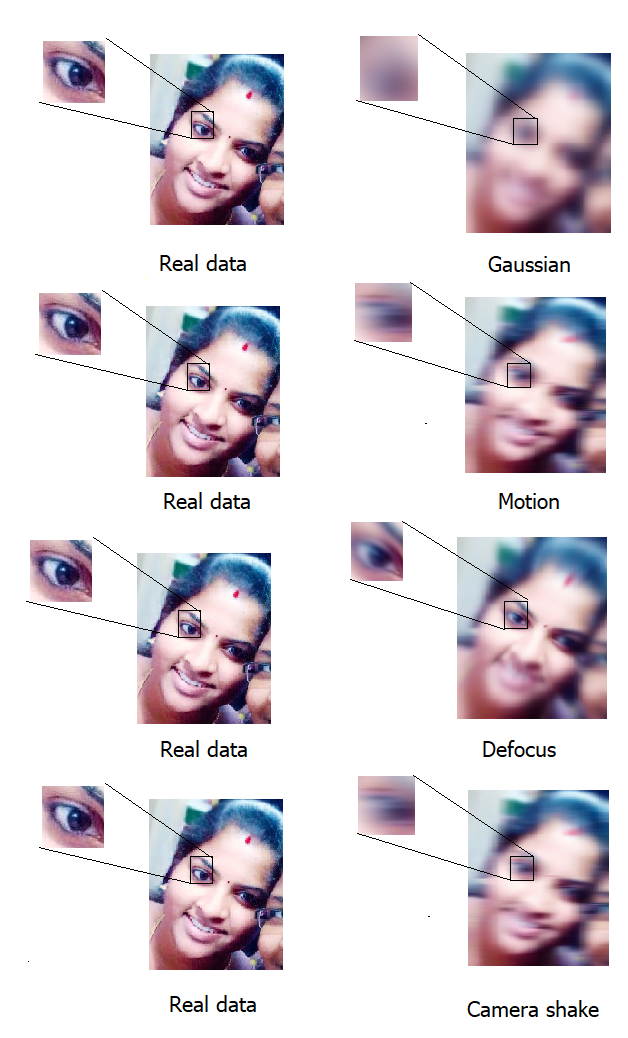} \label{dataset}
         \caption{Training data samples - Image pair}
\end{figure}

\subsection{Motion blur}

Motion blur is commonly occurred when capturing fast moving object and when long time of exposure is needed.  Motion blur caused bu the relative motion between the objects and sensing system.  When the object motions is relatively fast as compares  with the exposure period, the motion blur can occur as a linear motion blur. This can be represented by 1-D averaging of the neighbor pixels,

\[   
h(i, j, L, \theta) = 
     \begin{cases}
       \frac{1}{L} &\quad\text{if} \sqrt{(i^2 + j^2)} \leq \frac{L}{2} \text{and} \frac{i}{j} = - \tan \theta \\
       0 &\quad\text{Otherwise}\\
     \end{cases} \label{motion_blur}
\]
where, $(i,j)$ is the coordinates from the center $h$, $L$ is the moving distance and $\theta$ is the direction of the object moving.  Motion blur can be generated by the mentioned Equation. However, in real time,  the moving objects in the image only occupies a part of the image and the blur generated by such operation is extremely complicated. 

\subsection{camera shake blur}
Camera shake blur is another kind of blur which commonly occur in many real time cases. Unlike motion blur, it caused by the motion of the camera instead of the object which result with degradation of information.  Camera rotation cause the most complicated blur as it indulges the in-plane and our-of-plane rotation with respect to the focal plane. In in-plane rotation, kernel blur varies across the images from the camera rotation axis while in out-of-plane rotation, the degree of spatial variance across the image is dependent on the focal length of the camera. Also, it can occur in in low lighting conditions. The motion of the camera in an irregular directions cause the in plane or out of plane motion. While capturing a long distance object, light translation in the camera motion is spatially invariant. Hence, this can be modeled as linear blur motion,

\[   
h(i, j, L, \theta) = 
     \begin{cases}
       \frac{1}{L} &\quad\text{if} \sqrt{(i^2 + j^2)} \leq \frac{L}{2} \text{and} \frac{i}{j} = - \tan \theta \\
       0 &\quad\text{Otherwise}\\
     \end{cases} \label{motion_blur}
\]
During the camera translation, the objects nearer undergoes large shift, if different objects lies in different focal plane in a single scene that causes the large degree of information degradation in an image. 
\subsection{Defocus blur}
Defocus blur usually occurs at the image by improper focus of the image by the image sensing system. In different depths of scenes, the object outside the focus are highly suffocates from the defocus blur.  It may even occur by the single lens incorporation of the sensing system by acquiring information from the object out side of the depth of the focus. Defocus blur can be approximately modeled by uniform circular model,
\[   
h(i, j) = 
     \begin{cases}
       \frac{1}{\pi R^2} &\quad\text{if} \sqrt{(i^2 + j^2)} \leq R \\
       0 &\quad\text{Otherwise}\\
     \end{cases} \label{motion_blur}
\]where, $R$ is the radius of the circle.

\section{Result and Discussion}
The training of the GAN network to recover the image information is carried out by using NVDIA- GTX 980m GPU. Stochastic gradient descent with batch size of 4 and Adam optimizer are implemented to increase the learning speed and network convergence towards to global minimum. 

\begin{figure}[!hpb]
         \includegraphics[width=14cm]{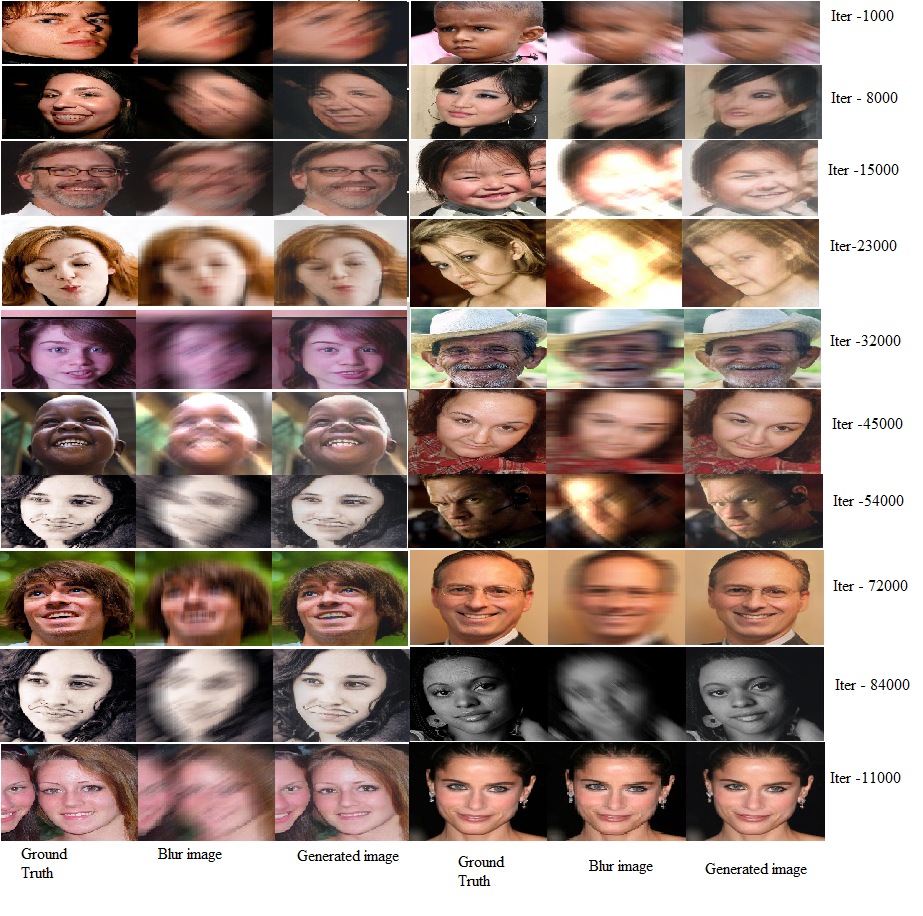} \label{validation}
         \caption{Generated images from artificially created blur images/ On training}
\end{figure}

\begin{figure}[!hpb]
         \includegraphics[width=12cm]{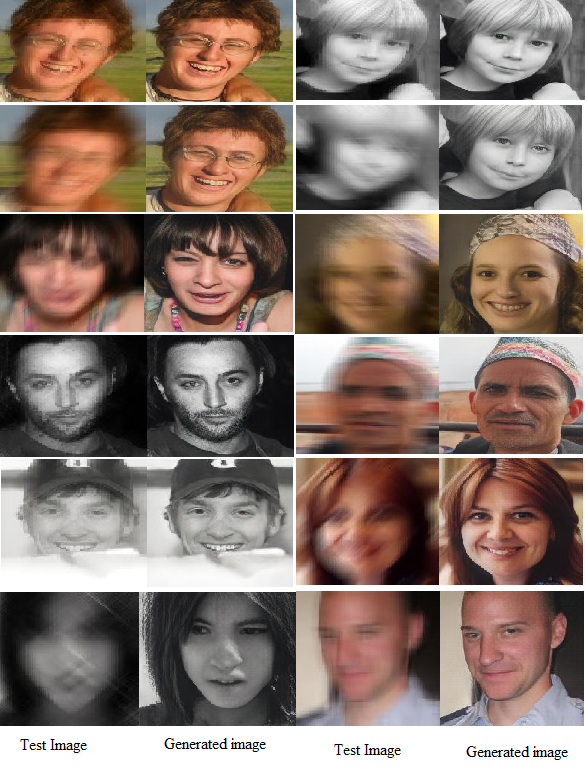} \label{test}
         \caption{Generated images from artificially created blur images/ On testing}
\end{figure}

The learning rate is set to $0.0002$ and $\beta_1$ $\&$ $\beta_2$ are set to $0.5 \& 0.999$ respectively.  The network is trained for 240 epochs for a week.  In order to acquire the probability density from the real data distribution, $256$ number of kernels with large pixel size of $7 \times 7$ is implemented in first and second convolutional layer in generator. 128 number of kernels are used in Resnet blocks with the kernel size of $5 \times 5$.  It is important to note that the Instance  BatchNorm is implemented in after every convolution operators to avoid discriminator loss quickly approach zero. LeakyReLU with $\alpha = 0.1$ is used as the non-linear activation function to avoid the sparse gradient. In up-sampling module,  PixelShuffle, Convolution Transpose with stride two is used while in down-sampling module, average Pooling, Convolution operation with stride of two is enchanted.  Without these up-sampling, down-sampling modules, generator suffers from the generation of undesirable pixel noise. Dropout with probability of $50 \%$ is used after the convolution operation.   To optimize the network, as mentioned in image to image translation GAN network, training is carried out to increase the $\log D(x_i, (G(x_i, z_i))$ instead minimizing the $\log [1 - D(x_i, (G(x_i, z_i))]$. Alternate gradient step is used between the generator and the discriminator.  The generated images on training is shown in the Figure \ref{validation}. First column of the images is the artificially created blurred images, middle column is the ground truth and the final column is the generated images. During the training, the network is generated more realistic images. Figure \ref{test} shows the generated images on test data. Left pair of images are the successful information on the testing. Right side pair images are the failure cases. During the testing of the architecture, many cases are resulted with failure. The network is capable of recover the information, if the network is trained on same family of degraded images. However, the network can be trained to have more generalization, if the network is trained on wide variety of image pairs.

\section{Conclusion}
A cGAN based framework for recovering the possible information from heavily blurred images is proposed. The training of the network is carried out by using adversarial loss and perceptional loss function.  The generated images from various experiments show that the addition of up-sampling and down-sampling module in the generator network is help to increase the performance of the network dramatically as well as to recover the informations. The primary objective is only to recover the information from the blurred image faces, only face images with blur is considered for training and testing. Hence, any comparative study between the state-of-art models isn't conducted. Also, another important conclusion is made from the experimentation that the network is capable of recover the information fully as long as the network is trained on the same verity of degraded images. This research work is carried out as a part of our IOP studio software development 'Facial recognition module'.    
 
\bibliographystyle{unsrt}
\bibliography{bmc_article}  

\end{document}